\documentclass[lettersize,journal]{IEEEtran}

\usepackage{balance}
\usepackage{color}
\usepackage{subcaption}
\usepackage{multirow}
\usepackage{array}
\usepackage{url}
\usepackage{hyperref}
\usepackage{amsmath}

\usepackage{booktabs}
\usepackage{multirow}
\usepackage{textcomp}
\usepackage{setspace}

\usepackage{amsmath,amssymb,amsfonts} 
\usepackage{algorithmic}
\usepackage{algorithm}
\usepackage{array}
\usepackage{textcomp}
\usepackage{stfloats}
\usepackage{verbatim}
\usepackage{graphicx}
\usepackage{graphics} 
\usepackage{bm}
\usepackage{multirow}
\usepackage{booktabs}
\usepackage[table,xcdraw]{xcolor}

\usepackage[backend=bibtex,natbib=true,style=numeric-comp,sorting=none,giveninits=true,maxbibnames=99,url=false,doi=false]{biblatex}
\begin{document}

\title{  
RIDERS: Radar-Infrared Depth Estimation \\ for Robust Sensing
}

\author{Han Li$^{1}$, Yukai Ma$^{1}$, Yuehao Huang$^{1}$, Yaqing Gu$^{1}$, Weihua Xu$^{1}$, Yong Liu$^{1,*}$, Xingxing Zuo$^{2,*}$
\thanks{$^{1}$The authors are with the Institute of Cyber-Systems and Control, Zhejiang University, Hangzhou, China.}
\thanks{$^{2}$ The author is with the Technical University of Munich, Munich, Germany.}%
\thanks{$^*$ Xingxing Zuo and Yong Liu are the corresponding authors (Email: {\tt\small xingxing.zuo@tum.de; yongliu@iipc.zju.edu.cn}).}
}

\markboth{IEEE TRANSACTIONS}%
{Li \MakeLowercase{\textit{et al.}}: RIDERS: Radar-Infrared Depth Estimation for Robust Sensing}

\maketitle

\setlength{\textfloatsep}{2.0pt} 

\begin{abstract}
Dense depth recovery is crucial in autonomous driving, serving as a foundational element for obstacle avoidance, 3D object detection, and local path planning. Adverse weather conditions, including haze, dust, rain, snow, and darkness, introduce significant challenges to accurate dense depth estimation, thereby posing substantial safety risks in autonomous driving. These challenges are particularly pronounced for traditional depth estimation methods that rely on short electromagnetic wave sensors, such as visible spectrum cameras and near-infrared LiDAR, due to their susceptibility to diffraction noise and occlusion in such environments.
To fundamentally overcome this issue, we present a novel approach for robust metric depth estimation by fusing a millimeter-wave Radar and a monocular infrared thermal camera, which are capable of penetrating atmospheric particles and unaffected by lighting conditions.
Our proposed Radar-Infrared fusion method achieves highly accurate and finely detailed dense depth estimation through three stages, including monocular depth prediction with global scale alignment, quasi-dense Radar augmentation by learning Radar-pixels correspondences, and local scale refinement of dense depth using a scale map learner. Our method achieves exceptional visual quality and accurate metric estimation by addressing the challenges of ambiguity and misalignment that arise from directly fusing multi-modal long-wave features. We evaluate the performance of our approach on the NTU4DRadLM dataset and our self-collected challenging ZJU-Multispectrum dataset. Especially noteworthy is the unprecedented robustness demonstrated by our proposed method in smoky scenarios. Our code will be released at \url{https://github.com/MMOCKING/RIDERS}.

\end{abstract}


\section{Introduction}

\IEEEPARstart{P}{erception} plays a critical role in autonomous driving and robotics, with depth estimation serving as the preliminary for dense reconstruction, obstacle avoidance, and 3D detection. 
Vehicles equipped with advanced driver assistance systems commonly utilize LiDAR, which operates in the near-infrared spectrum, and RGB cameras that capture the visible spectrum. This sensor fusion provides a comprehensive perception of complex 3D environments, combining accurate geometric range data with detailed visual information. Such integration is regarded as a dependable approach in various driving contexts.

\begin{figure}[t]
      \centering
      \includegraphics[width=0.49\textwidth]{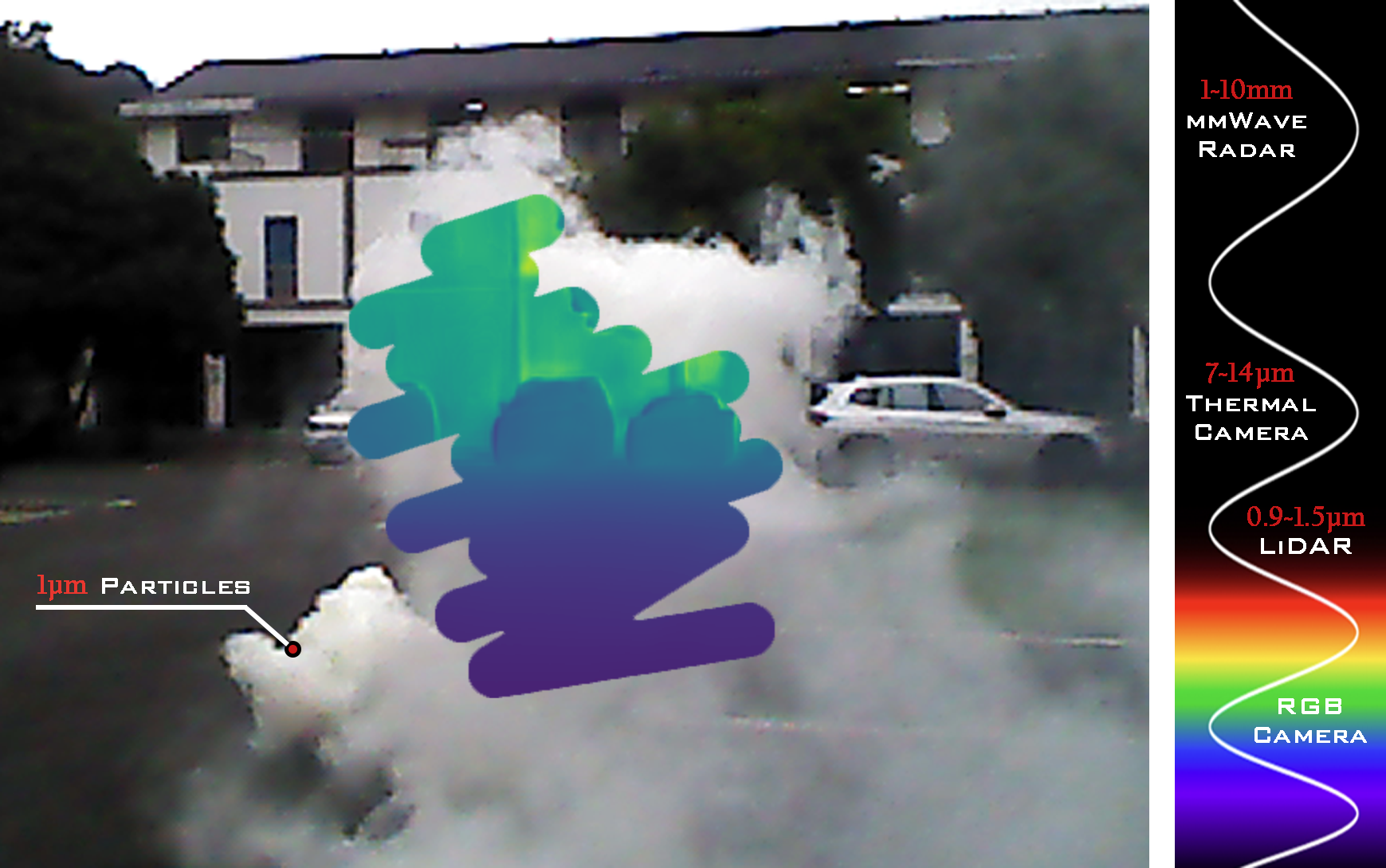}
      \captionsetup{font=small}
      \caption{Left: Our approach can provide high-quality depth estimation beyond the visible spectrum, unaffected by micrometer-sized particles. Right: Millimeter-wave Radar and infrared thermal cameras have longer operational wavelengths than LiDAR and RGB cameras to penetrate atmospheric particles.}
      \label{fig:fm}
      \vspace{5pt} 
\end{figure}

However, due to the inherent limitations of LiDAR and RGB cameras, this combination can fail concurrently in adverse weather conditions like haze, dust, smoke, fog, rain, or snow. The visible light wavelength ranges from 390nm to 780nm, while mainstream LiDAR emits laser wavelengths of 905nm and 1550nm \cite{rablau2019lidar}. In the presence of micrometer-sized atmospheric particles, serious diffraction occurs, severely interfering with the imaging or echo reception. 
Existing solutions that enhance perceptual robustness at the software level \cite{zhao2020robust, liu2021self, gasperini_morbitzer2023md4all} often face challenges in overcoming the inherent limitations of sensors. Methods that fit well on datasets or are effective against mild disturbances struggle to maintain performance in complex and dynamic real-world driving scenarios. Even the best denoising methods may fail in scenes with complete occlusion. Therefore, there is a demand for a robust depth estimation method that can fundamentally handle interferences such as smoke, rain, and snow. Addressing this challenge requires a perspective shift toward robust sensor solutions.

In the field of autonomous driving, mmWave Radars and infrared thermal cameras, known for their high resilience in adverse weather conditions, are increasingly recognized. Automotive imaging Radars, primarily operating at 77GHz within the W-band \cite{tessmann2002compact}, feature a wavelength of approximately 3.9mm. In contrast, thermal cameras generally operate within a wavelength range of 7-14$\mu$m. Due to their minimal susceptibility to atmospheric particulates and variable lighting conditions, these sensors offer a viable alternative to the conventional LiDAR and RGB camera ensemble, especially in challenging environments such as nighttime, smoke-filled, or foggy conditions.

Current methodologies integrating Radar and infrared camera technology predominantly focus on target detection and pose estimation tasks \cite{zhangu2021traffic, doer2021radar, zhang20234drt}. However, the potential for their fusion in achieving 3D dense depth estimation remains largely untapped. Similar to the limitations of monocular solutions with RGB cameras, depth estimation using a single infrared camera struggles to accurately gauge the metric scale of a 3D scene \cite{kim2018multispectral, lu2021alternative, shin2021self, shin2022maximizing, shin2023self}. In contrast, combining Radar and infrared sensors allows for delivering metric dense depth. 
Although the fusion of Radar with RGB cameras for depth estimation has demonstrated considerable success \cite{lin2020depth, long2021radar, lo2021depth, gasperini2021r4dyn, singh2023depth}, the direct integration of either extracted features or raw data from infrared imagery with millimeter-wave Radar data poses substantial challenges. These difficulties stem from the inherently low contrast and lack of textures in infrared images, combined with Radar data's inherently noisy and sparse nature.
Severe aliasing, blurring, and artifacts can compromise the quality of the resulting dense depth maps.

\begin{figure*}[t]
      \centering
      \includegraphics[width=0.99\textwidth]{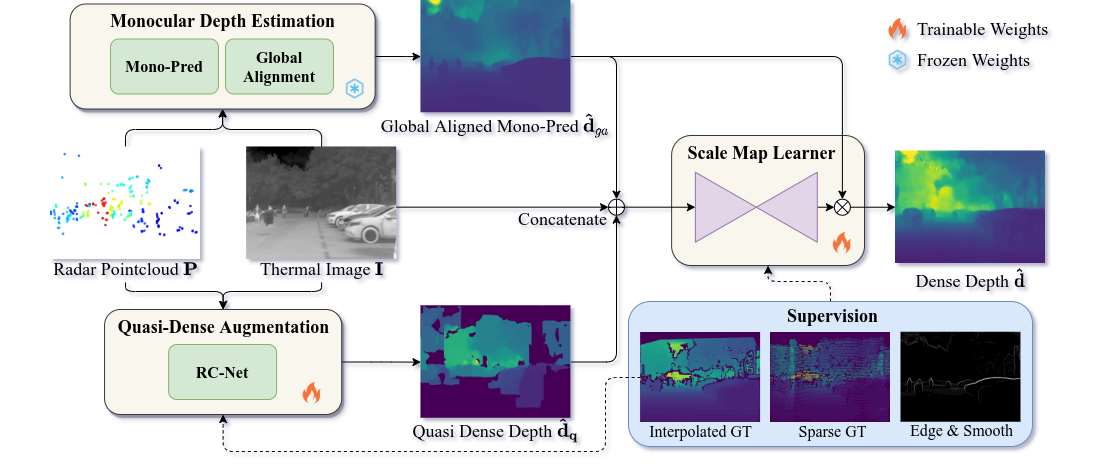}
      \caption{The overall framework of our proposed RIDERS is comprised of three stages: monocular depth estimation from infrared images, quasi-dense augmentation of Radar depth, and scale map learner for refining the local scale of dense depth.}
      \label{fig:pipeline}
      \vspace{0pt} 
\end{figure*}

To overcome these challenges, this paper introduces RIDERS, a novel approach for dense depth estimation through the fusion of Radar and thermal camera data. This work extends our previous conference paper RadarCam-Depth~\cite{li2024radarcam}, which focused on combining Radar with an RGB camera for depth estimation. Adopting a similar underlying philosophy, RIDERS is tailored to integrate Radar and thermal camera data effectively, particularly in environments with smoke, fog, or low light.
To the best of our knowledge, our work represents the \textbf{\textit{first}} attempt to integrate an infrared camera and a Radar for metric dense depth estimation. 
This synergistic sensor fusion yields significant advantages: (i) The integration of a thermal camera, which captures detailed relative dense depth, with Radar, which provides metric scales for sampled points, enables the generation of metric dense depth maps with enhanced detail. (ii) Operating in the mid-infrared and millimeter-wave spectrums, these sensors effectively circumvent most atmospheric particulates, ensuring perception characterized by low diffraction and minimal occlusion. (iii) Immune to ambient light variations, thermal imaging, and millimeter-wave reflections facilitate consistent environmental perception, regardless of the time of day.

Specifically, we achieve the dense metric depth estimation from a Radar and a thermal camera with three stages. Firstly, we employ an off-the-shelf generalizable monocular prediction model to obtain scaleless depth from the thermal images and perform global alignment with sparse Radar point clouds. Concurrently, we learn the confidence of association between Radar points and pixels in thermal images, enhancing sparse Radar points into quasi-dense depth. Ultimately, we utilize the scale map learner to adjust local scales for monocular estimation results and recover the final dense depth. Our method tries to fully leverage the advantages of multi-modal data at each component stage. 
This three-stage paradigm boasts two significant advantages: (i) We circumvent the direct fusion of raw data or encodings of heterogeneous point clouds and images, thereby preventing aliasing artifacts and preserving high-fidelity fine details in dense depth estimation (see Fig.~\ref{fig:fm}). (ii) Unlike direct depth estimation with a wide convergence basin, our approach essentially involves recovering dense scale for scale-free monocular depth, leveraging the strong prior of mono-depth and achieving higher learning efficiency.

The primary contributions of this work can be summarized as follows:
\begin{itemize}
    \item Presenting the \textbf{\textit{first}} known dense depth estimation approach that integrates mmWave Radar and thermal cameras, possessing unparalleled robustness for depth perception in adverse conditions such as smoke and low lighting.
    \item Introducing a novel metric dense depth estimation framework that effectively fuses heterogeneous Radar and thermal data. Our three-stage framework comprises monocular estimation and global alignment, quasi-dense Radar augmentation, and dense scale learning, ultimately recovering dense depth from sparse and noisy long-wave data.
    \item The proposed method has undergone extensive testing on the publicly available NTU dataset \cite{zhang2023ntu4dradlm} and the self-collected ZJU-Multispectrum dataset, surpassing other solutions and demonstrating high metric accuracy and solid robustness.
    \item Our high-quality ZJU-Multispectrum dataset containing challenging scenarios with 4D Radar, thermal camera, RGB camera data, and reference depth from 3D LiDAR will be released. We will also open-source our code at \url{https://github.com/MMOCKING/RIDERS} to fertilize future research. 
\end{itemize}

\section{Related Works} \label{sec:related_works}

\subsection{Depth from Monocular Infrared Thermal Image}

The infrared spectrum band exhibits high-level robustness against adverse weather and lighting conditions. However, infrared images lack texture information compared to visible spectrum images, appearing more blurred and suffering from a scarcity of large-scale datasets.
Consequently, numerous existing methods attempt to transfer knowledge from the visible spectrum to thermal depth estimation tasks. 
Kim et al.'s Multispectral Transfer Network (MTN) \cite{kim2018multispectral} is trained with chromaticity clues from RGB images, enabling stable depth prediction from monocular thermal images.
Lu et al. \cite{lu2021alternative} propose using a CycleGAN-based generator to transform RGB images into fake thermal images, creating a stereo pair of a thermal camera for supervising the disparity prediction.
Shin et al.'s approach \cite{shin2021self} leverages multispectral consistency for self-supervised depth estimation, incorporating temperature consistency from thermal imaging and photometric consistency from wrapped RGB images.

However, the above methods require closely matching and often RGB and thermal images from identical scenes and viewpoints, imposing stringent data requirements.
Recently, Shin et al. \cite{shin2023self, shin2023joint} proposed a method that does not require paired multispectral data. Their network consists of modality-specific feature extractors and
modality-independent decoders. They train the network to achieve feature-level adversarial adaptation, minimizing the gap between RGB and thermal features.
ThermalMonoDepth \cite{shin2022maximizing} is a self-supervised depth estimation method that eliminates the need for extra RGB involvement in training. It introduces a time-consistent image mapping method reorganizing thermal radiation values and ensuring temporal consistency, maximizing self-supervision for thermal image depth estimation. 
Additionally, Shin et al. \cite{shin2023deep} propose a unified depth network that effectively bridges monocular thermal depth and stereo thermal depth tasks from a conditional random field approach perspective. 
Nevertheless, monocular methods often lack accurate scale and are prone to local optima in self-supervised training, leading to poor metric accuracy.

\subsection{Depth from Radar-Camera Fusion}

The fusion of Radar and RGB camera data for metric depth estimation is an active area of research. Lin et al. \cite{lin2020depth} introduced a two-stage CNN-based pipeline that combines Radar and camera inputs to denoise Radar signals and estimate dense depth. Long et al. \cite{long2021full} proposed a Radar-2-Pixel (R2P) network, utilizing radial Doppler velocity and induced optical flow from images to associate Radar points with corresponding pixel regions, enabling the synthesis of full-velocity information. They also achieved image-guided depth completion using Radar and video data \cite{long2021radar}.
Another approach, DORN \cite{lo2021depth}, proposed by Lo et al., extends Radar points in the elevation dimension and applies a deep ordinal regression network-based \cite{fu2018deep} feature fusion. Unlike other methods, R4dyn \cite{gasperini2021r4dyn} creatively incorporates Radar as a weakly supervised signal into a self-supervised framework and employs Radar as an additional input to enhance robustness. However, their method primarily focuses on vehicle targets and does not fully correlate all Radar points with a larger image area, resulting in lower depth accuracy.
Nevertheless, the methods above typically require multi-frame information to overcome the sparsity of Radar data. In contrast, Singh et al. \cite{singh2023depth} presented a fusion method that relies solely on a single image frame and Radar point cloud. Their first-stage network infers the confidence scores of associating a Radar point to image patches, leading to a semi-dense depth map after association. They further employ a gated fusion network to control the fusion of multi-modal Radar-Camera data and predict the final dense depth.

These methods directly encode the multi-modal inputs and learn the target depth. However, direct encoding and concatenation of the inherently ambiguous Radar depth and images can confuse the learning pipeline, resulting in aliasing and other undesirable artifacts in the estimated depth. 
In our preceding conference version~\cite{li2024radarcam}, we explored scale learning for monocular depth estimation using RGB imagery supplemented by Radar data. However, given the RGB camera's sensitivity to lighting conditions, there is a pressing need for a robust method of metric dense depth estimation using infrared cameras, which are unaffected by variations in ambient illumination. Compared to \cite{li2024radarcam}, we have enhanced the performance of Radar depth augmentation by adjusting the quasi-dense depth calculation strategy. Additionally, the introduction of original thermal images as input, along with the incorporation of smoothness loss, has improved the performance of our Scale Map Learner.

\begin{figure*}[t]
      \centering
      \includegraphics[width=0.99\textwidth]{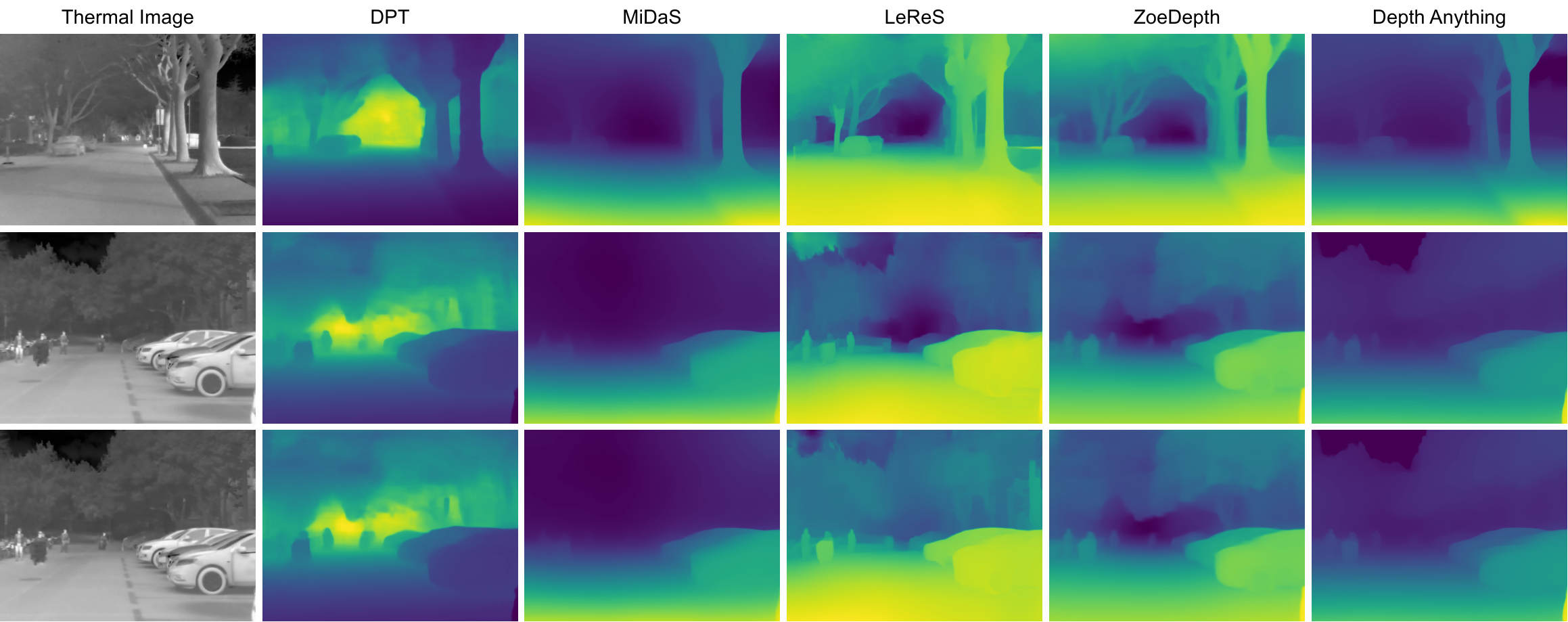}
      \caption{\textbf{Zero-shot generalized depth predictions}. From left to right: the input thermal image, zero-shot generalized depth prediction from DPT \cite{ranftl2021vision}, MiDaS \cite{birkl2023midas}, ZoeDepth \cite{bhat2023zoedepth}, LeReS \cite{yin2021learning,yin2022towards}, and Depth Anything \cite{yang2024depth}, which are trained on RGB images. 
      The second and third rows correspond to consecutive frames with a time interval of about 0.1 seconds.
      LeReS, ZoeDepth, and Depth Anything exhibit good performance without fine-tuning. Specifically, LeReS provides precise edges and fine details, while ZoeDepth and Depth Anything demonstrate temporal consistency in consecutive frames.}
      \label{fig:leres}
      \vspace{0pt} 
\end{figure*}

\section{Methodology}
\label{sec:methodology}

Our goal is to recover the dense depth $\mathbf{\hat{d}} \in \mathbb{R}^{H_0 \times W_0}_+$ from a pair of thermal image $\mathbf{I} \in \mathbb{R}^{C\times H_0\times W_0}$ and Radar point cloud $\mathbf{P} = \left\{\mathbf{p}_i | \mathbf{p}_i \in \mathbb R^3, i = 0,1,2,\cdots,k-1\right\}$ transformed into the thermal camera coordinate through known extrinsic calibration. 
As different datasets may use either single-channel thermal images or three-channel pseudo-color images, the number of channels $C$ for the image $\mathbf{I}$ can be either 1 or 3.
The overall framework of our RIDERS consists of three main stages: monocular depth prediction and alignment (Sec.~\ref{mono depth}), quasi-dense augmentation of sparse Radar (Sec.~\ref{RC-Net}), and scale map learner (SML) for refining dense scale locally (Sec.~\ref{SML}), as shown in Fig.~\ref{fig:pipeline}.

\subsection{Monocular Depth Prediction and Scale Alignment} 
\label{mono depth}

\subsubsection{Monocular Depth Prediction}

This module provides prior monocular depth $\mathbf{\hat{d}}_{m}$ for dense scale learning from a single-view thermal image. $\mathbf{\hat{d}}_{m}$ does not require high metric accuracy but needs to provide relative depth reflecting the high-fidelity image details, for example, object edges and surface smoothness. 

Contrary to the abundance of large-scale datasets for RGB images in the visible spectrum, there is a notable scarcity of equivalent datasets for training monocular depth prediction models specifically for infrared imagery. Owing to the analogous imaging principles—both being passive receivers of electromagnetic waves—and the comparable modality between thermal and RGB images, we can utilize a monocular depth prediction model, originally trained on RGB images, directly on thermal images. This zero-shot generalization approach results in preliminary sub-optimal dense depth for monocular thermal images. 

With abundant datasets for monocular depth prediction for RGB images, monocular depth prediction designed for high generalization has gained increasingly robust performance through extensive training \cite{ranftl2020towards, yin2021learning, ranftl2021vision, yin2022towards, birkl2023midas, bhat2023zoedepth, yang2024depth}. Existing methods rely on the depth reconstruction loss in the scale- and shift-invariant space on multiple datasets, enabling the recovery of high-quality scale-invariant relative depth. These models exhibit powerful zero-shot generalization capabilities, even when applied to monochrome thermal images. We can obtain the preliminary monocular depth $\mathbf{\hat{d}}_m$ (or inverse depth $\mathbf{\hat{z}}_m$) from the input $\mathbf{I}$ using any monocular depth prediction network. In other words, our module in this section utilizes a replaceable network, which can be continuously updated with the advancement of the monocular depth prediction networks. 
In the context of generalization on thermal images, pre-trained models like LeReS \cite{yin2021learning, yin2022towards}, ZoeDepth \cite{bhat2023zoedepth}, and Depth Anything \cite{yang2024depth} demonstrate reasonable accuracy. 
LeReS provides dense depth maps with clear object edges and structures, though it may exhibit depth discontinuities between adjacent frames. On the other hand, ZoeDepth and Depth Anything achieve high temporal consistency in their depth estimates, with Depth Anything particularly distinguished by its superior ability to differentiate background elements, such as the sky. (See Fig.~\ref{fig:leres}).

\subsubsection{Global Scale Alignment}
\label{scale estimator}

In order to enhance the efficiency of refining pixel-wise scale by SML in the proceeding stage, we align the scale-free monocular depth prediction $\mathbf{\hat{d}}_m$ with the depth of Radar points $\mathbf{P}$ using a global scaling factor $\hat{s}_g$, which generates globally aligned depth $\mathbf{\hat{d}}_{ga}$. 
Empirically, our findings suggest that a single scaling factor is adequate for global alignment, diverging from existing methodologies that employ both scaling and shifting factors for this purpose \cite{li2024radarcam, naumann2023nerf}.

To optimize the monocular depth estimate $\mathbf{\hat{d}}_m$, we employ the bounded Brent numerical optimization algorithm~\cite{forsythe1977computer, brent2013algorithms}. Our optimization objective is to find the optimal scaling factor $\hat{s}_g$ that minimizes the following loss function:
\begin{align}
    \mathcal{F}(\hat{s}_g, \mathbf{\hat{d}}_m, d(\mathbf{P}), \mathbf{M_P}) = \mathbf{M_P} \cdot | \hat{s}_g \cdot \mathbf{\hat{d}}_m - d(\mathbf{P}) |,
\end{align}
where $\hat{s}_g$ is the scaling factor to be optimized, $d(\cdot)$ returns the depth of Radar points in image coordinates, and $\mathbf{M_P} \in \left\{0,1\right\}^{H\times W}$ is the valid position mask of Radar points on the image. We define Radar points with depths in the range of 0-100m as valid ones.

We utilize empirical values to set initial bounds for the optimization target $\hat{s}_g$. Ultimately, the Brent algorithm employs a combination of quadratic interpolation, the secant method, and bisection to compute the optimal solution within the predefined bounds iteratively.
The globally aligned metric monocular depth is then calculated as $\mathbf{\hat{d}}_{ga} = \hat{s}_g \cdot \mathbf{\hat{d}}_m$. Its inverse depth $\mathbf{\hat{z}}_{ga}$ is subsequently fed into the scale map learner (SML).

\begin{figure*}[t]
      \centering
      \includegraphics[width=0.97\textwidth]{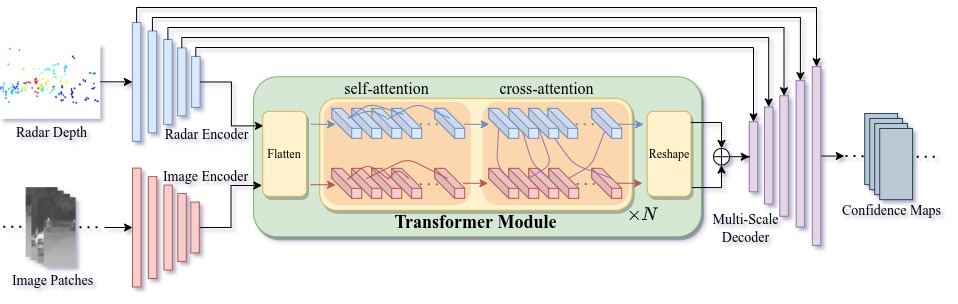}
      \caption{\textbf{The architecture of our sparse Radar augmentation network, RC-Net.} The network takes Radar depths and image patches cropped around each Radar point as input. The architecture consists of two encoder branches, a transformer module, and a multi-scale decoder, aiming to infer pixel-level confidence scores for each Radar point-image patch pairing.}
      \label{fig:rcnet}
      \vspace{0pt} 
\end{figure*}

\subsection{Quasi-Dense Radar Augmentation} 
\label{RC-Net}
Due to inherent sparsity and noises in Radar data, additional augmentation of Radar depth is crucial before conducting scale learning.
To densify the sparse Radar depth obtained from projection, we exploit a transformer-based Radar-Camera data association network (shorthand RC-Net), which predicts the confidence of Radar-pixel associations. 
The proposed network predicts the association confidence between pixels in $\mathbf{I}$ and neighboring Radar points. It utilizes a weighted average of the depths from multiple Radar points to calculate the depth for each pixel. Ultimately, this module outputs a quasi-dense map with continuous depth, denoted as $\mathbf{\hat{d}}_q$.

\subsubsection{Network Architecture}
Our RC-Net (refer to Fig.~\ref{fig:rcnet}) is derived from the vanilla RC-vNet \cite{singh2023depth} and is enhanced by integrating self and cross-attention mechanisms \cite{sun2021loftr} within a transformer module.
The image encoder of RC-Net adopts a standard ResNet backbone~\cite{he2016deep}, while the Radar encoder comprises a multi-layer perceptron with fully connected layers. We reproject the sparse Radar points into the image plane to generate a sparse Radar depth map, which is input into the Radar encoder. The extracted Radar features undergo mean pooling and are reshaped to match the width and height of image features.
Subsequently, Radar and image features are flattened and processed through $N=4$ layers of self and cross-attention.
To be specific, the attention layer takes inputs—query $\mathbf{Q}$, key $\mathbf{K}$, and value $\mathbf{V}$. It calculates weights by assessing the similarity between $\mathbf{Q}$ and $\mathbf{K}$, converts similarity values to weights via a softmax layer, and produces the attention value by weighting and summing the values, as shown in the following equation:
\begin{align}
    {\rm{Attention}}(\mathbf{Q},\mathbf{K},\mathbf{V})={\rm{softmax}}(\mathbf{Q} \mathbf{K}^\top)\mathbf{V}.
\end{align}

In our task, when $\mathbf{Q}$ and $\mathbf{V}$ are features from the same modality, the attention mechanism is self-attention. When $\mathbf{Q}$ and $\mathbf{V}$ come from Radar and image features, respectively, the attention mechanism is cross-attention. This mechanism allows for better estimating the correlation between different elements, providing our multi-modal fusion with a larger receptive field.

Ultimately, the features encoded through the transformer module, along with skip connections from the middle layers of the encoder, are input into the decoder. The output is in logit form, and the final step involves activating the logits through the sigmoid function, resulting in confidence maps for cross-modal associations.

\subsubsection{Confidence of Cross-Modal Associations}
\begin{figure}[h]
    \centering
    \begin{subfigure}[t]{0.22\textwidth}
        \centering
        \tiny
        \includegraphics[width=\columnwidth]{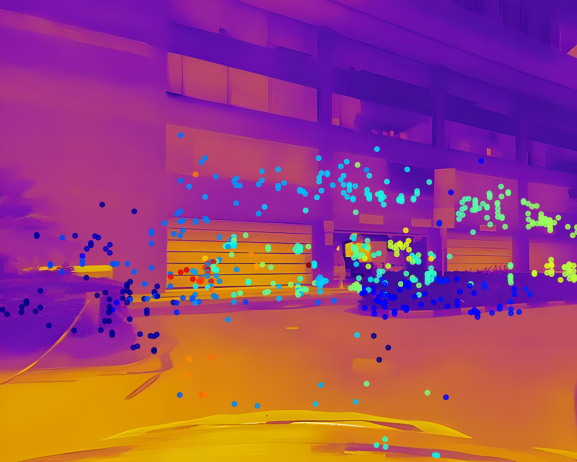}
    \end{subfigure}
    \centering
    \begin{subfigure}[t]{0.22\textwidth}
        \centering
        \tiny
        \includegraphics[width=\columnwidth]{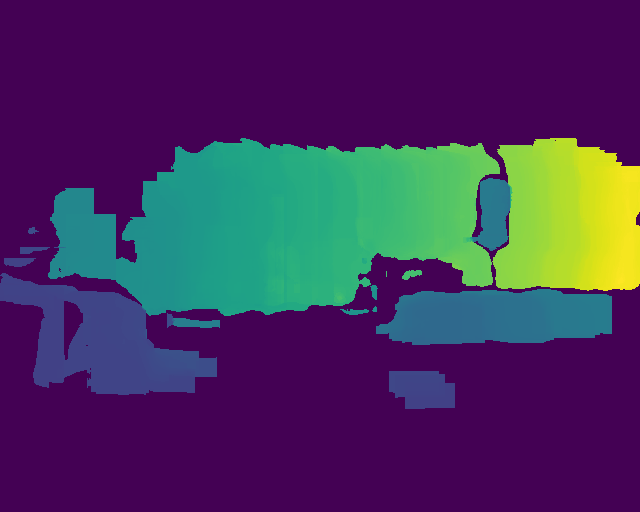}
    \end{subfigure}

    \captionsetup{font=small}
    \caption{\textbf{Radar augmentation result.} Left: sparse Radar points back-projected onto the thermal image plane. Right: augmented quasi-dense depth $\hat{\mathbf{d}}_q$ from our RC-Net.
    }
    \label{fig:dq} 
\end{figure}

For a Radar point $\mathbf{p}_i$ and a cropped image patch $\mathbf{Z}_i \in \mathbb{R}^{C\times H\times W}$ in its projection vicinity, we utilize RC-Net $h_{\theta}$ to produce a confidence map $\mathbf{\hat{y}}_i=h_{\theta}(\mathbf{Z}_i, \mathbf{p}_i)\in [0,1]^{H\times W}$, representing the probability of whether the pixels in $\mathbf{Z}_i$ corresponds to $\mathbf{p}_i$, inspired by~\cite{singh2023depth}. With all $k$ points in a Radar point cloud $\mathbf{P}$, the forward pass generates $k$ confidence maps for individual Radar points.
Therefore, each pixel $\mathbf{x}_{uv}$ within $\mathbf{I}$ $(u\in[0,W_0-1], v\in[0,H_0-1])$ has $n \in [0,k]$ associated Radar point candidates. 
By selecting confidence scores above the threshold, we can identify potential associated Radar points $\mathbf{P}_\mu$ for pixel $\mathbf{x}_{uv}$. Then we compute the depth of pixel $\mathbf{x}_{uv}$ by taking a weighted average of all $\mathbf{P}_\mu$ depths using their normalized confidence scores as weights, resulting in a quasi-dense depth map $\hat{\mathbf{d}}_q$ as shown in Fig.~\ref{fig:dq}.
\begin{align}
    \hat{\mathbf{d}}_q(u,v) =
    \begin{cases} 
    d(\mathbf{P}_\mu) \cdot \mathbf{\hat{y}}_\mu(x_{uv})/\Sigma\mathbf{\hat{{y}}}_\mu(x_{uv}),\ \text{if}\ \mu \neq \emptyset, \\
    \text{None}, \ \ \ \ \ \ \ \ \ \ \ \ \ \ \ \ \ \ \ \ \ \ \ \ \ \ \text{otherwise},
    \end{cases}
\end{align}
where $\mu = \{ i\ |\ \mathbf{\hat{y}}_i(x_{uv})>\tau \}$, $\mathbf{P}_{\mu}$ is the corresponding Radar point set, $\mathbf{\hat{{y}}}_\mu(x_{uv})$ is the corresponding confidence score, $d(\cdot)$ returns the depth value. Finally, quasi-dense scale map $\mathbf{\hat{s}}_q$ is calculated from $\mathbf{\hat{s}}_q=\mathbf{\hat{d}}_{q} / \mathbf{\hat{d}}_{ga}$, and its inverse $1/\mathbf{\hat{s}}_q$ is subsequently fed into the scale map learner.

\subsubsection{Training}
We begin by projecting LiDAR point clouds to obtain $\mathbf{d}_{gt}$ in the image coordinate. Subsequently, we perform linear interpolation in log space \cite{barber1996quickhull} on $\mathbf{d}_{gt}$, resulting in $\mathbf{d}_{int}$. 
For supervision, we use $\mathbf{d}_{int}$ to create binary classification labels $\mathbf{y}_{i} \in \left\{0,1\right\}^{H\times W}$, where $\mathbf{d}_{int}$ pixels with a depth value deviation less than 0.5m from the Radar point are labeled as positive.
Following the construction of $\mathbf{y}_{i}$, we minimize the binary cross-entropy loss:
\begin{align}
  \begin{split}
    \mathcal{L}_{BCE}=\frac{1}{|\Omega|}\sum_{x\in \Omega}-(\mathbf{y}_{i}(x)\log \mathbf{\hat{y}}_i(x)\\
    +(1-\mathbf{y}_{i}(x))\log(1-\mathbf{\hat{y}}_i(x))),
    \end{split}
\end{align}
where $\Omega \subset \mathbb{R}^2$ denotes the image region of $\mathbf{Z}_i$, $x \in \Omega$ is a pixel coordinate, and $\mathbf{\hat{y}}_i = h_{\theta}(\mathbf{Z}_i,\mathbf{p}_i)$ is the confidence of correspondence.

\subsection{Scale Map Learner}
\label{SML}
\subsubsection{Network Architecture}
Drawing inspiration from \cite{wofk2023monocular}, we construct a scale map learner (SML) network based on MiDaS-small \cite{ranftl2020towards} architecture. SML aims to learn a pixel-wise dense scale map for $\mathbf{\hat{z}}_{ga}$, thereby completing the quasi-dense scale map and refining the metric accuracy of $\mathbf{\hat{z}}_{ga}$. SML requires concatenated $\mathbf{I}$,  $\mathbf{\hat{z}}_{ga}$ and $1/\mathbf{\hat{s}}_q$ as input. The empty parts in $\mathbf{\hat{s}}_q$ are filled with ones. SML regresses a dense scale residual map $\mathbf{r}$, where values can be negative. The final scale map is derived as $1/\mathbf{\hat{s}} = \text{ReLU}(1 + \mathbf{r})$, and the ultimate metric depth estimation is computed as $\mathbf{\hat{d}} = \mathbf{\hat{s}} /\mathbf{\hat{z}}_{ga}$.

\subsubsection{Training}
During training, ground truth depth $\mathbf{d}_{gt}$ is derived from the projection of 3D LiDAR points. Linear interpolation \cite{barber1996quickhull} in log space is further performed to get a densified $\mathbf{d}_{int}$. We minimize the difference between the estimated metric depth $\hat{\mathbf{d}}$ and the ground truth $\mathbf{d}_{gt}$ and $\mathbf{d}_{int}$ with a L1 penalty:
\begin{align}
    \mathcal{L}_{depth}=\mathcal{L}(\mathbf{d}_{int},\hat{\mathbf{d}})+\lambda_{gt}\mathcal{L}(\mathbf{d}_{gt},\hat{\mathbf{d}}),
\end{align}
\begin{align}
    \mathcal{L}(\mathbf{d},\hat{\mathbf{d}})=
    \frac{1}{|\Omega_d|}\underset{x\in \Omega_d}{\sum} |\mathbf{d}(x)-\hat{\mathbf{d}}(x)|,
\end{align}
where $\lambda_{gt}$ is the weight of $\mathcal{L}_{gt}$, $\Omega_{d} \subset \Omega$ denotes the domains where ground truth has valid depth values. 

In addition, we incorporate a smoothness loss constraint. Leveraging the fact that $\mathbf{\hat{d}}_{ga}$ provides depth rather than texture-based edges, we exploit this to encourage smoothness in the non-edge regions of the output $\mathbf{\hat{d}}$, thereby enhancing the overall estimation quality.
We employ Sobel filters \cite{vincent2009descriptive} to compute the gradients $\triangledown x_{ga}$ and $\triangledown y_{ga}$ of $\mathbf{\hat{d}}_{ga}$, as well as the gradients $\triangledown x$ and $\triangledown y$ of $\mathbf{\hat{d}}$. This process allows us to formulate the smoothness loss $\mathcal{L}_{smooth}$:
\begin{align}
    \mathcal{L}_{smooth} = e^{-|\triangledown x_{ga}|} \cdot |\triangledown x| + e^{-|\triangledown y_{ga}|} \cdot |\triangledown y|.
\end{align}

Our final loss is expressed as:

\begin{align}
    \mathcal{L}_{SML} = \mathcal{L}_{depth} + \lambda_{smooth} \mathcal{L}_{smooth},
\end{align}
where $\lambda_{smooth}$ is the weight of $\mathcal{L}_{smooth}$.

\section{Experiments}
\label{section:EXPERIMENTS}

\subsection{Datasets}
\subsubsection{NTU4DRadLM Dataset}
We first evaluate the proposed method on the NTU4DRadLM dataset~\cite{zhang2023ntu4dradlm}. Designed explicitly for SLAM and sensing research, this dataset integrates 4D Radar, thermal cameras, and IMUs and comprises approximately 30,000 synchronized LiDAR-Radar-Thermal keyframes. For testing, we selected the sequences ``loop2-2022-06-03-1'' and ``loop3-2022-06-03-0'', which collectively consist of 6,039 frames. The remaining 24,160 frames served as our training and validation splits. However, it is pertinent to note that the publicly accessible portion of the NTU4DRadLM dataset predominantly features typical scenes under clear weather conditions, limiting the scope for robustness assessment.

\begin{figure}[h]
      \centering
      \includegraphics[width=0.49\textwidth]{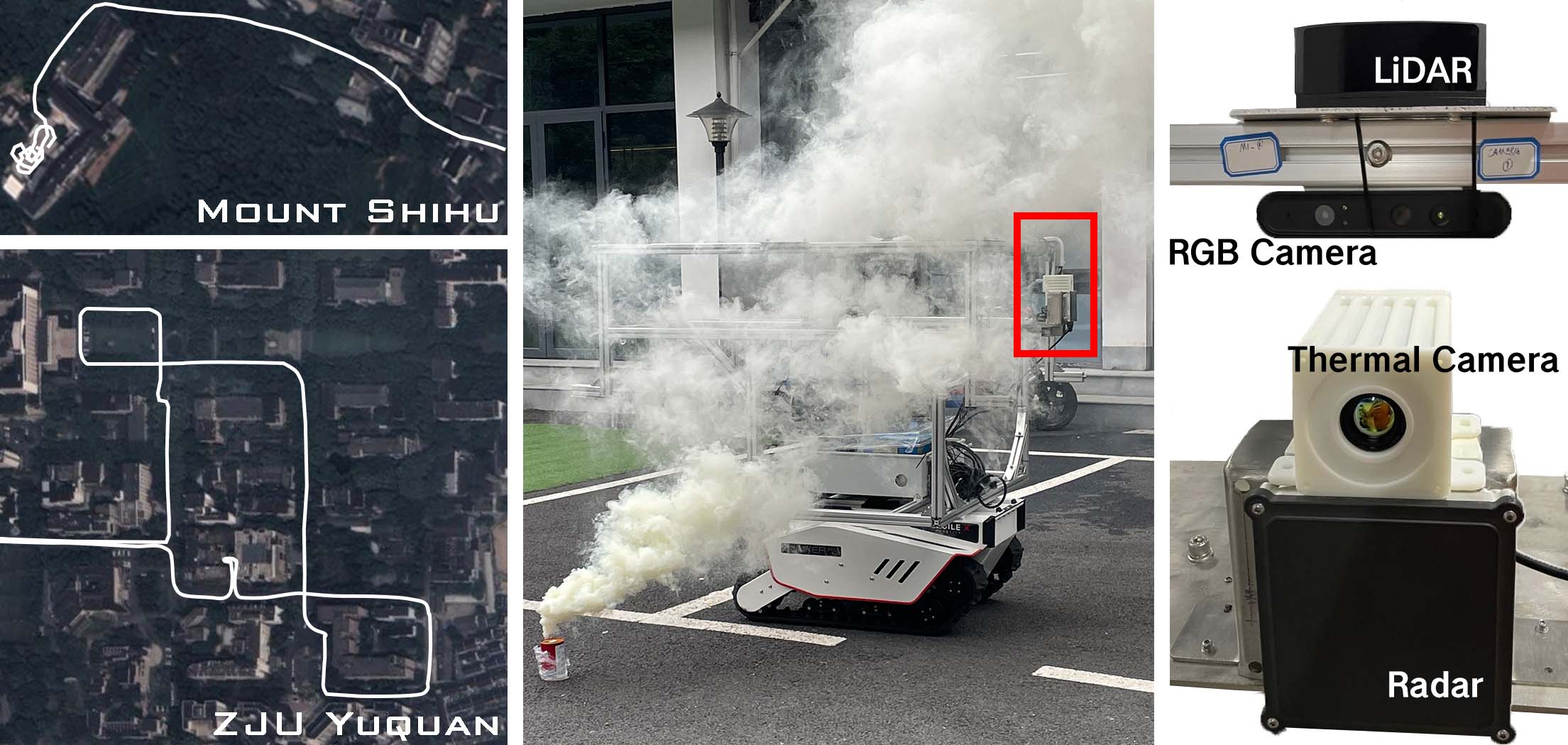}
      \captionsetup{font=small}
      \caption{\textbf{ZJU-Multispectrum dataset.} From left to right: data collection path of ZJU-Multispectrum in Mount Shihu and Zhejiang University's Yuquan Campus, our vehicle equipped with multi-modal sensors, and the sensors mounted on the vehicle.}
      \label{fig:sensors}
      \vspace{0pt} 
\end{figure}

\subsubsection{ZJU-Multispectrum Dataset}
To assess the robustness of the proposed method in challenging conditions, such as smoke and nighttime environments, we collect data using our customized ground robot platform (see Fig.~\ref{fig:sensors}). Our data collection vehicle is equipped with the following sensors: Geometrical-pal 4D imaging Radar, RoboSense M1 LiDAR, Dali VD641 thermal infrared camera, and Orbbec Astra RGBD camera. Utilizing these sensors, we collected data across various scenarios, encompassing daytime, nighttime, clear weather, and conditions with artificial smoke generated by smoke canisters. The training and validation set involved sequences captured in clear weather, consisting of eight daytime sequences and one nighttime sequence, 21,522 synchronized frames in total.
For quantitative evaluation in daytime scenarios, we selected three clear-weather sequences comprising a total of 5,532 frames. To assess performance in more challenging conditions, we utilized three additional sequences. Specifically, we examined the method's resilience to low-light environments through a nighttime clear-weather sequence of 184 frames. Furthermore, we evaluated the method's effectiveness in smoke conditions with one daytime and one nighttime smoke sequence, together providing LiDAR depth ground truth for 446 frames. It should be noted that the sensors were maintained in a stationary position for the 446 frames with depth ground truth. This was essential to ensure the acquisition of LiDAR depth of the surroundings, which served as the reference for our evaluation. The remaining frames of two smoke sequences, with sensors moving in the floating and diffusing smoke, are used for the visualization showcase.

\begin{table*}[th]
\centering
\caption{\textsc{Evaluations on NTU4DRadLM} (mm).}
\label{NTU test}
\resizebox{0.95\linewidth}{!}{
\begin{tabular}{c|c|c|c|c|c|c|c|c}
\hline\hline
{\textbf{Range}} & \textbf{Method} & \textbf{iMAE} $\downarrow$ & \textbf{iRMSE} $\downarrow$ & \textbf{MAE} $\downarrow$ & \textbf{RMSE} $\downarrow$ & \textbf{AbsRel} $\downarrow$& \textbf{SqRel} $\downarrow$ & $\mathbf{\delta_1}$ $\uparrow$ \\

\hline
\multirow{6}*{0-50m} 
& DORN \cite{lo2021depth} & 5.903 & 10.575 & 1915.141 & \textbf{4458.021} & 0.090 & \textbf{806.773} & 0.920 \\
{} & Singh \cite{singh2023depth} & 5.340 & 10.256 & 1866.829 & 4586.367 & 0.092 & 1092.425 & 0.915 \\
{} & RacarCam-Depth \cite{li2024radarcam} & 5.065 & 10.173 & 2004.375 & 4869.808 & 0.094 & 1157.161 & 0.908 \\
{} & \textbf{RIDERS (LeReS)} & 4.585 & 9.539 & 1785.232 & 4552.015 & 0.084 & 1029.812 & 0.921 \\
{} & \textbf{RIDERS (ZoeDepth)} & \textbf{4.377} & 9.459 & \textbf{1745.794} & 4552.851 & \textbf{0.082} & 1053.554 & 0.923 \\
{} & \textbf{RIDERS (Depth Anything)} & 4.506 & \textbf{9.445} & 1760.916 & 4492.681 & 0.083 & 1017.192 & \textbf{0.924} \\

\hline
\multirow{6}*{0-60m} 
& DORN \cite{lo2021depth} & 6.040 & 10.664 & 2323.629 & 5329.139 & 0.097 & \textbf{979.009} & 0.908 \\
{} & Singh \cite{singh2023depth} & 5.307 & 10.270 & 2120.584 & 5076.233 & 0.094 & 1154.673 & 0.909 \\
{} & RacarCam-Depth \cite{li2024radarcam} & 5.091 & 10.293 & 2321.156 & 5493.308 & 0.097 & 1246.593 & 0.900 \\
{} & \textbf{RIDERS (LeReS)} & 4.590 & 9.633 & 2040.465 & 5056.552 & 0.086 & 1094.782 & 0.915 \\
{} & \textbf{RIDERS (ZoeDepth)} & \textbf{4.381} & 9.544 & \textbf{1988.879} & 5027.688 & \textbf{0.084} & 1111.248 & \textbf{0.918}\\
{} & \textbf{RIDERS (Depth Anything)} & 4.504 & \textbf{9.525} & 2003.297 & \textbf{4967.266} & 0.085 & 1074.892 & 0.918\\

\hline
\multirow{6}*{0-70m} 
& DORN \cite{lo2021depth} & 6.112 & 10.706 & 2622.396 & 6013.591 & 0.101 & \textbf{1096.660} & 0.900 \\
{} & Singh \cite{singh2023depth} & 5.288 & 10.289 & 2344.816 & 5577.079 & 0.096 & 1222.454 & 0.904 \\
{} & RacarCam-Depth \cite{li2024radarcam} & 5.122 & 10.427 & 2619.444 & 6152.787 & 0.100 & 1349.044 & 0.892 \\
{} & \textbf{RIDERS (LeReS)} & 4.602 & 9.713 & 2283.267 & 5606.387 & 0.088 & 1171.620 & 0.909 \\
{} & \textbf{RIDERS (ZoeDepth)} & \textbf{4.392} & 9.619 & \textbf{2217.410} & 5538.507 & \textbf{0.086} & 1179.719 & 0.913 \\
{} & \textbf{RIDERS (Depth Anything)} & 4.515 & \textbf{9.602} & 2236.068 & \textbf{5498.191} & 0.087 & 1146.673 & \textbf{0.913} \\

\hline\hline
\end{tabular}
}
\vspace{0pt} 
\end{table*}

\subsection{Training Details and Evaluation Protocol}

For both two datasets, LiDAR point clouds projected onto the image plane were used as the ground truth $\mathbf{d}_{gt}$ for training and evaluation. Subsequently, linear interpolation \cite{barber1996quickhull} was performed on $\mathbf{d}_{gt}$ in the logarithmic space of depth, yielding $\mathbf{d}_{int}$ as the dense supervisory signal.

For training the RC-Net for quasi-dense Radar augmentation, we adopted the network architecture of Sec.~\ref{RC-Net}.
When training on NTU4DRadLM, with an input image size of $512 \times 640$, the size of the cropped patch for confidence map generation in RC-Net is set to $150\times50$. For ZJU-Multispectrum, the input image size is $480 \times 640$, and the patch size is $240\times100$. 
The training employed the Adam optimizer, with $\beta_1 =0.9$ and $\beta_2 =0.999$, and a learning rate of $2e^{-4}$ for 100 epochs. Data augmentations, including horizontal flipping, saturation, brightness, and contrast adjustments, are applied with a probability of 0.5. 

We employ the MiDaS-Small network architecture for our Scale Map Learner (SML). The encoder backbone is initialized with pre-trained ImageNet weights \cite{deng2009imagenet}, and other layers are randomly initialized. The input data is resized to a fixed height of $288$ and a width that is a multiple of $32$. We use an Adam optimizer with $\beta_1=0.9$ and $\beta_2=0.999$. The initial learning rate is set to $1e^{-4}$ and reduced to $5e^{-5}$ after 20 epochs. Data augmentations, including horizontal flipping and random Radar noise, are employed.
All training and testing activities were carried out on a single RTX 3090 GPU.

Some widely adopted metrics from the literature are used for evaluating the depth estimations, including mean absolute error (MAE), root mean squared error (RMSE), absolute relative error (AbsRel), squared relative error (SqRel), the errors of inverse depth (iRMSE, iMAE), and $\delta_1$~\cite{sun2021neuralrecon}. 
To better illustrate our experimental details, we have prepared a demo video, which is available for viewing at \url{https://youtu.be/wRsRTZoWUpE}.

\begin{figure}[h]
      \centering
      \includegraphics[width=0.485\textwidth]{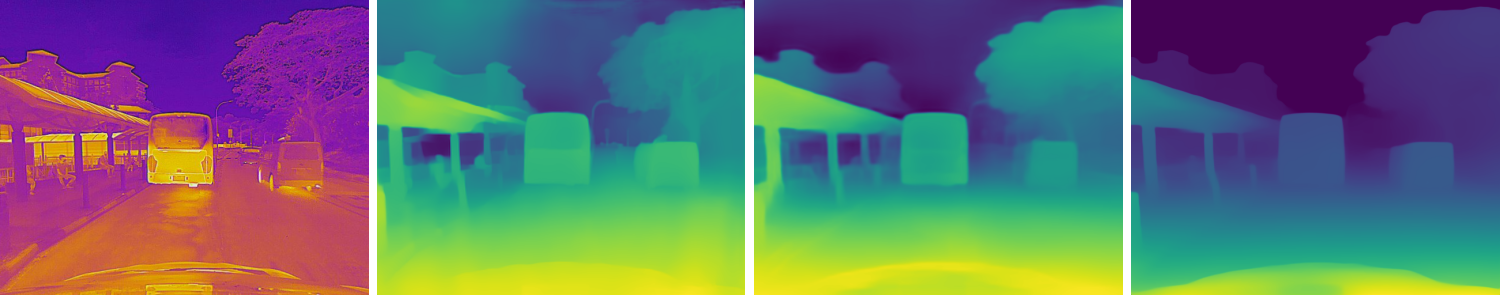}
      \captionsetup{font=small}
      \caption{\textbf{Mono-Preds of the NTU4DRadLM images.} From left to right: the input thermal image, scale-free inv-depth $\mathbf{\hat{z}}_{m}$ from LeReS \cite{yin2021learning}, ZoeDepth \cite{bhat2023zoedepth} and Depth Anything \cite{yang2024depth}.}
      \label{fig:ntu-mono}
      \vspace{0pt} 
\end{figure}

\begin{figure*}[t]
      \centering
      \includegraphics[width=1\textwidth]{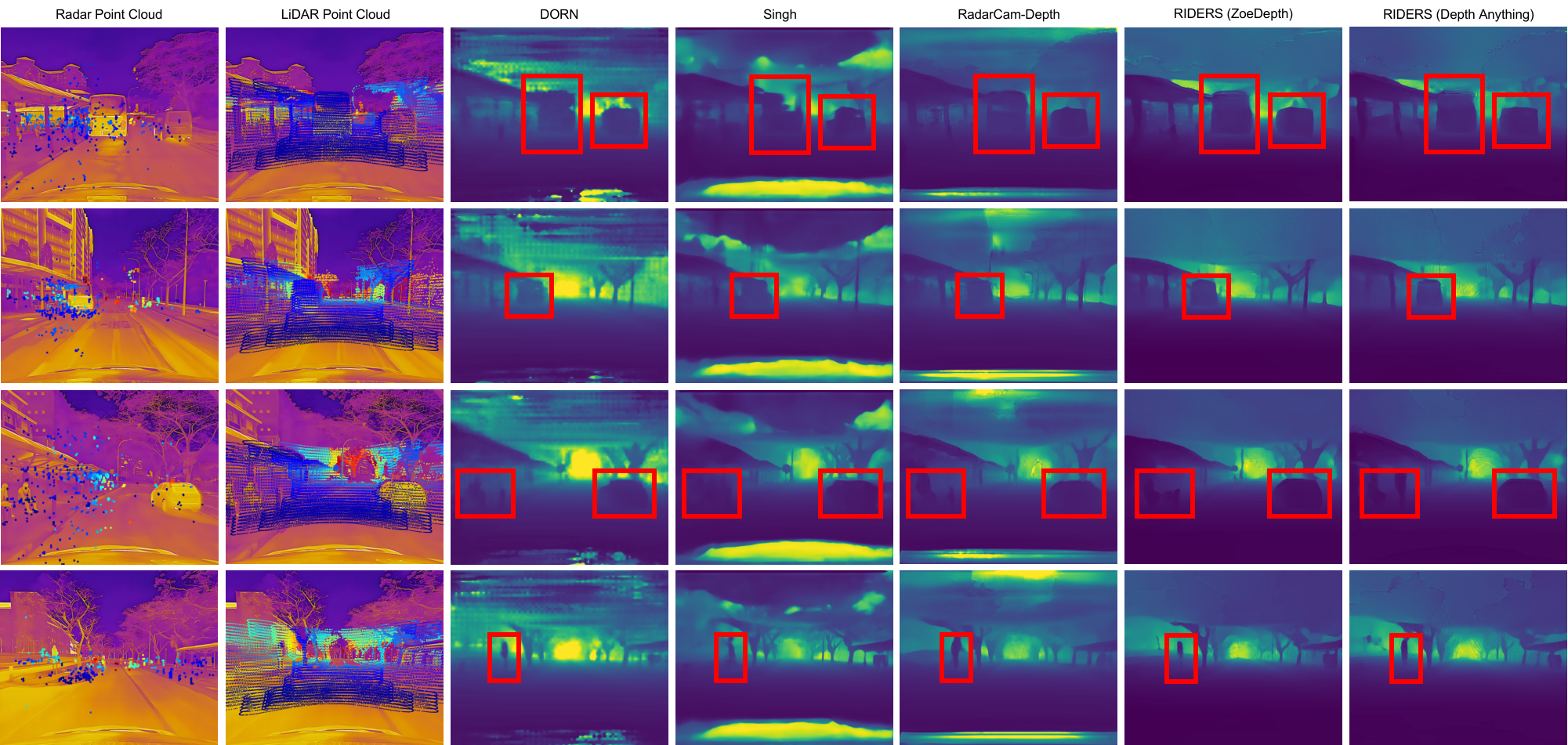}
      \caption{\textbf{Evaluations on NTU4DRadLM.} From left to right: Raw Radar points $\mathbf{P}$ overlaid on the infrared image, LiDAR point clouds $\mathbf{\hat{d}}_{gt}$ overlaid on the infrared image,  and depth predictions from $\mathbf{\hat{d}}$ of DORN \cite{lo2021depth}, Singh \cite{singh2023depth}, and our RIDERS based on monocular depth prediction models, ZoeDepth \cite{bhat2023zoedepth} and Depth Anything \cite{yang2024depth}, on NTU4DRadLM dataset~\cite{zhang2023ntu4dradlm}. Harnessing the preliminary monocular depth prediction, our method demonstrates superior performance. Specifically, for crucial objects in traffic scenes, such as vehicles and pedestrians, our approach provides clear outlines of the targets (highlighted by red boxes in the images). [The monocular depth predictions from the pre-trained ZoeDepth and Depth Anything models, corresponding to the first row, are illustrated in Fig.~\ref{fig:ntu-mono}.]}
      \label{fig:compare_ntu}
      \vspace{0pt} 
\end{figure*}

\begin{figure*}[t]
      \centering
      \includegraphics[width=1\textwidth]{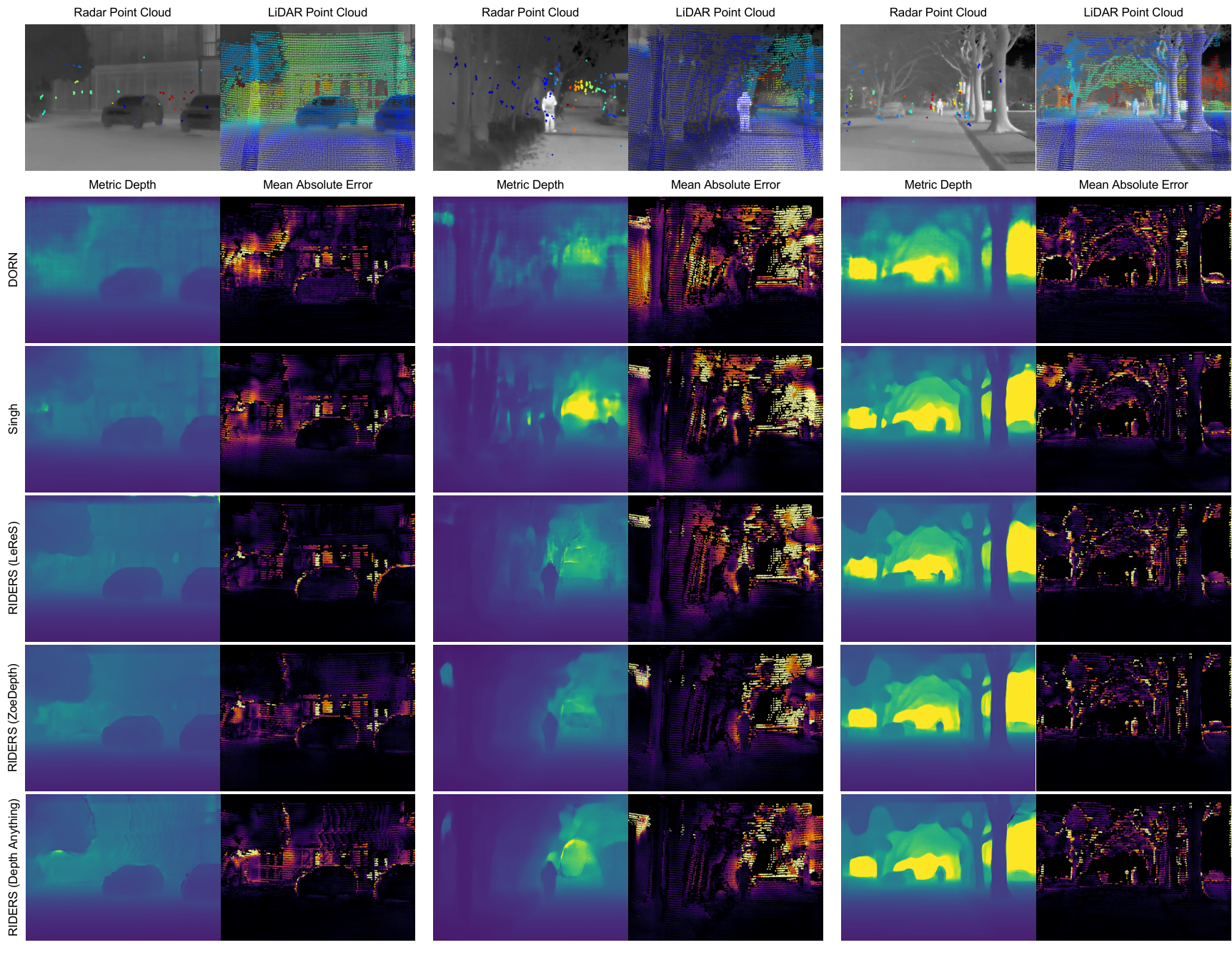}
      \caption{\textbf{Evaluations on ZJU-Multispectrumc (clear-day).} First row: Radar (left) and LiDAR (right) point cloud of ZJU-Multispectrum dataset. The remaining five rows, from top to bottom, show the depth estimation results (left) of DORN \cite{lo2021depth}, Singh \cite{singh2023depth}, and our RIDERS ($\mathbf{\hat{d}}_m$ from LeRes \cite{yin2021learning} and ZoeDepth \cite{bhat2023zoedepth}) along with error visualizations (right).}
      \label{fig:compare_zju_error}
      \vspace{0pt} 
\end{figure*}

\subsection{Evaluation on NTU4DRadLM}
\label{evaluation on ntu}
This section evaluates the metric dense depth against $\mathbf{d}_{gt}$ within the range of 50, 60, and 70 meters on the NTU4DRadLM dataset. Due to the lack of existing methods based on Radar-Infrared camera fusion, we compare with methods \cite{lo2021depth,singh2023depth,li2024radarcam} originally designed for Radar and RGB camera fusion. All methods were thoroughly trained and validated on the thermal images and Radar point cloud data from the dataset. We also tested the RIDERS method using different monocular depth estimation modules, including LeReS \cite{yin2021learning}, ZoeDepth \cite{bhat2023zoedepth}, and Depth Anything \cite{yang2024depth}.

Compared to RIDERS with monocular estimates from LeReS, the RIDERS using ZoeDepth and Depth Anything exhibit superior metric accuracy. This is attributed to the excellent consistency in their predictions, resulting in fewer depth discontinuities between consecutive frames. This characteristic is advantageous for precise scale learning in the SML module.
On the other hand, our method outperforms other Radar-Camera fusion methods on most evaluation metrics (see Tab.~\ref{NTU test}). Within the evaluation ranges of 50m, 60m, and 70m, RIDERS (ZoeDepth) shows a reduction of 13.6\%, 13.9\%, and 14.3\% in iMAE compared to the second-best method, respectively. Similarly, AbsRel is reduced by 8.9\%, 10.6\%, and 10.4\%, respectively. We attribute the outstanding performance of RIDERS to the reasonable monocular depth prediction  $\mathbf{\hat{d}}_{m}$ (see Fig.~\ref{fig:ntu-mono}) and our scale learning strategy. RIDERS exhibits excellent qualitative results regarding the edges and morphology of objects such as vehicles and buildings, as shown in Fig.~\ref{fig:compare_ntu}. Scale learning based on $\mathbf{\hat{d}}_{ga}$ outputs avoids issues like blurring and deformation that may arise in direct-depth metric learning methods \cite{lo2021depth,singh2023depth}. 
Compared to our previous work, RadarCam-Depth \cite{li2024radarcam} developed for RGB images, RIDERS has effectively improved the adaptability and accuracy for low-contrast and blurry thermal images. The augmented Radar depth calculation in our RC-Net greatly enhances the metric accuracy (see Sec.~\ref{sec:radar depth aug}). Furthermore, the introduction of raw thermal imagery as an input to the SML provides an auxiliary source of information, thereby refining the predictions derived solely from $\mathbf{\hat{d}}_{ga}$.

\begin{table*}[th]
\centering
\caption{\textsc{Evaluations on ZJU-Multispectrum (clear-day)} (mm).}
\label{ZJU clear test}
\resizebox{0.95\linewidth}{!}{
\begin{tabular}{c|c|c|c|c|c|c|c|c}
\hline\hline
{\textbf{Range}} & \textbf{Method} & \textbf{iMAE} $\downarrow$ & \textbf{iRMSE} $\downarrow$ & \textbf{MAE} $\downarrow$ & \textbf{RMSE} $\downarrow$ & \textbf{AbsRel} $\downarrow$& \textbf{SqRel} $\downarrow$ & $\mathbf{\delta_1}$ $\uparrow$ \\

\hline
\multirow{8}*{0-50m} 
 & DORN \cite{lo2021depth} & 11.184 & 19.176 & 2975.231 & 5376.028 & 0.162 & 1256.015 & 0.786 \\
{} & Singh \cite{singh2023depth} & 11.282 & 17.386 & 2979.062 & 5579.791 & 0.164 & 1547.239 & 0.788 \\
{} & RacarCam-Depth \cite{li2024radarcam} & 10.647 & 16.484 & 2917.770 & 5433.830 & 0.167 & 1611.953 & 0.798 \\

\cline{2-9}
{} & Direct Depth (ZoeDepth) & 11.763 & 17.231 & 2923.381 & 4822.382 & 0.170 & 1175.563 & 0.763 \\
{} & Sparse Input (ZoeDepth) & 10.035 & 15.149 & 2721.244 & 5019.822 & 0.148 & 1219.950 & 0.806 \\

\cline{2-9}
{} & \textbf{RIDERS (LeReS)} & 10.423 & 15.755 & 2658.893 & 4765.552 & 0.147 & 1084.529 & 0.806 \\
{} & \textbf{RIDERS (ZoeDepth)} & 9.413 & 14.291 & 2576.012 & 4755.345 & 0.137 & 1081.774 & 0.826 \\
{} & \textbf{RIDERS (Depth Anything)} & \textbf{8.495} & \textbf{13.796} & \textbf{2423.499} & \textbf{4587.857} & \textbf{0.127} & \textbf{1025.969} & \textbf{0.848}\\

\hline
\multirow{8}*{0-60m} 
 & DORN \cite{lo2021depth} & 11.255 & 20.526 & 3213.565 & 5983.413 & 0.166 & 1376.385 & 0.779 \\
{} & Singh \cite{singh2023depth} & 11.292 & 17.420 & 3247.038 & 6099.891 & 0.167 & 1664.067 & 0.781 \\
{} & RacarCam-Depth \cite{li2024radarcam} & 10.587 & 16.433 & 3115.653 & 5772.308 & 0.168 & 1668.136 & 0.793 \\

\cline{2-9}
{} & Direct Depth (ZoeDepth) & 11.745 & 17.201 & 3213.202 & 5422.726 & 0.173 & 1288.683 & 0.753\\
{} & Sparse Input (ZoeDepth) & 10.011 & 15.130 & 2947.856 & 5483.022 & 0.150 & 1303.773 & 0.798\\

\cline{2-9}
{} & \textbf{RIDERS (LeReS)} & 10.443 & 15.788 & 2939.058 & 5363.344 & 0.150 & 1202.025 & 0.797\\
{} & \textbf{RIDERS (ZoeDepth)} & 9.435 & 14.341 & 2830.197 & 5297.403 & 0.140 & 1188.412 & 0.818 \\
{} & \textbf{RIDERS (Depth Anything)} & \textbf{8.510} & \textbf{13.822} & \textbf{2655.641} & \textbf{5076.447} & \textbf{0.130} & \textbf{1114.202} & \textbf{0.840}\\

\hline
\multirow{8}*{0-70m} 
 & DORN \cite{lo2021depth} & 11.287 & 20.546 & 3345.805 & 6394.996 & 0.167 & 1453.923 & 0.775\\
{} & Singh \cite{singh2023depth} & 11.298 & 17.444 & 3384.523 & 6407.994 & 0.168 & 1726.581 & 0.777 \\
{} & RacarCam-Depth \cite{li2024radarcam} & 10.575 & 16.422 & 3231.842 & 6023.848 & 0.169 & 1711.887 & 0.791 \\

\cline{2-9}
{} & Direct Depth (ZoeDepth) & 11.772 & 17.220 & 3419.844 & 5945.049 & 0.175 & 1390.536 & 0.747\\
{} & Sparse Input (ZoeDepth) & 10.029 & 15.155 & 3112.811 & 5878.983 & 0.152 & 1380.064 & 0.794\\

\cline{2-9}
{} & \textbf{RIDERS (LeReS)} & 10.482 & 15.835 & 3128.167 & 5843.109 & 0.152 & 1297.107 & 0.792\\
{} & \textbf{RIDERS (ZoeDepth)} & 9.461 & 14.379 & 2992.761 & 5699.685 & 0.142 & 1266.466 & 0.814 \\
{} & \textbf{RIDERS (Depth Anything)} & \textbf{8.532} & \textbf{13.852} & \textbf{2806.616} & \textbf{5449.972} & \textbf{0.131} & \textbf{1182.343} & \textbf{0.836}\\

\hline\hline
\end{tabular}
}
\vspace{0pt} 
\end{table*}

\begin{table*}[th]
\centering
\caption{\textsc{Evaluations on ZJU-Multispectrum (Night/Smoke)} (mm).}
\label{ZJU smoke test}
\resizebox{0.95\linewidth}{!}{
\begin{tabular}{c|c|c|c|c|c|c|c|c}
\hline\hline
\textbf{Sequence} & \textbf{Method} & \textbf{iMAE} $\downarrow$ & \textbf{iRMSE} $\downarrow$ & \textbf{MAE} $\downarrow$ & \textbf{RMSE} $\downarrow$ & \textbf{AbsRel} $\downarrow$ & \textbf{SqRel} $\downarrow$ & $\mathbf{\delta_1}$ $\uparrow$ \\

\hline
\multirow{6}*{Clear-Day}
& DORN \cite{lo2021depth} & 7.626 & 10.707 & 2498.583 & 4001.246 & 0.122 & 699.280 & 0.838 \\
{} & Singh \cite{singh2023depth} & 9.592 & 14.019 & 2955.268 & 4755.387 & 0.148 & 1105.797 & 0.781\\
{} & RacarCam-Depth \cite{li2024radarcam} & 9.558 & 13.431 & 2983.785 & 4748.783 & 0.159 & 1115.849 & 0.782\\

{} & \textbf{RIDERS (LeReS)} & 7.987 & 11.428 & 2473.828 & 4005.093 & 0.122 & 701.728 & 0.835\\
{} & \textbf{RIDERS (ZoeDepth)} & \textbf{5.973} & \textbf{8.730} & 2081.379 & 3519.996 & \textbf{0.098} & 526.128 & 0.897 \\
{} & \textbf{RIDERS (Depth Anything)} & 6.123 & 8.903 & \textbf{2006.024} & \textbf{3441.091} & 0.099 & \textbf{499.517} & \textbf{0.899} \\

\hline
\multirow{6}*{Clear-Night} 
& DORN \cite{lo2021depth} & 9.117 & 11.994 & 2273.446 & 3825.647 & 0.121 & 645.963 & 0.794 \\
{} & Singh \cite{singh2023depth} & 8.986 & 11.796 & 2228.139 & 3852.676 & 0.124 & 653.333 & 0.832\\
{} & RacarCam-Depth \cite{li2024radarcam} & 8.119 & 11.081 & 2166.027 & 3785.561 & 0.125 & 673.940 & 0.847 \\

{} & \textbf{RIDERS (LeReS)} & 8.244 & 11.309 & 2108.396 & 3950.610 & 0.109 & 621.577 & 0.848\\
{} & \textbf{RIDERS (ZoeDepth)} & 7.873 & 10.796 & 2082.876 & 3785.653 & 0.109 & 598.189 & 0.832\\
{} & \textbf{RIDERS (Depth Anything)} & \textbf{6.752} & \textbf{9.242} & \textbf{1705.355} & \textbf{3198.993} & \textbf{0.093} & \textbf{440.929} & \textbf{0.895}\\

\hline
\multirow{6}*{Smoke-Night} 
& DORN \cite{lo2021depth} & 3.947 & 5.941 & 1300.994 & 2522.735 & 0.067 & 272.588 & 0.954 \\
{} & Singh \cite{singh2023depth} & 3.900 & 6.589 & 1273.503 & 2697.400 & 0.066 & 299.175 & 0.952 \\
{} & RacarCam-Depth \cite{li2024radarcam} & 4.948 & 7.692 & 1836.872 & 3269.745 & 0.095 & 496.367 & 0.915 \\

{} & \textbf{RIDERS (LeReS)} & \textbf{3.143} & \textbf{5.129} & 1123.309 & 2430.145 & \textbf{0.054} & 219.441 & 0.965\\
{} & \textbf{RIDERS (ZoeDepth)} & 3.635 & 5.633 & \textbf{1110.664} & \textbf{2273.283} & 0.056 & \textbf{188.195} & \textbf{0.969}\\
{} & \textbf{RIDERS (Depth Anything)} & 3.530 & 5.848 & 1175.626 & 2402.630 & 0.057 & 215.350 & 0.962\\

\hline
\multirow{6}*{Smoke-Day} 
& DORN \cite{lo2021depth} & 6.901 & 9.991 & 4424.448 & 7120.837 & 0.149 & 1472.820 & 0.788 \\
{} & Singh \cite{singh2023depth} & 5.870 & 8.712 & 4539.365 & 7798.484 & 0.149 & 1796.066 & 0.792 \\
{} & RacarCam-Depth \cite{li2024radarcam} & 5.736 & 6.792 & 4048.047 & 6615.171 & 0.139 & 1322.224 & 0.839 \\

{} & \textbf{RIDERS (LeReS)} & \textbf{4.047} & 5.545 & \textbf{2702.623} & 4091.246 & \textbf{0.092} & 513.222 & 0.901\\
{} & \textbf{RIDERS (ZoeDepth)} & 4.182 & \textbf{5.039} & 2812.817 & 4711.455 & 0.095 & 616.680 & 0.924 \\
{} & \textbf{RIDERS (Depth Anything)} & 5.318 & 6.801 & 2765.294 & \textbf{3866.913} & 0.105 & \textbf{486.345} & \textbf{0.926}\\

\hline\hline
\end{tabular}
}
\vspace{0pt} 
\end{table*}

\subsection{Evaluation on ZJU-Multispectrum}
\label{evaluation on zju}
\subsubsection{Clear Daytime} 
We follow a similar way to Sec.~\ref{evaluation on ntu} for the evaluations on the ZJU-Multispectrum dataset. 
We first evaluate RIDERS on three clear daytime sequences. Our RIDERS outperformed DORN \cite{lo2021depth}, Singh \cite{singh2023depth} and RadarCam-Depth \cite{li2024radarcam} across all metrics, as shown in Tab.~\ref{ZJU clear test} and Fig.~\ref{fig:compare_zju_error}. 
Regarding monocular depth estimation modules, Depth Anything is the best choice.
Within the evaluation ranges of 50m, 60m, and 70m, RIDERS (Depth Anything) exhibits a reduction in iMAE compared to the second-best method by 20.2\%, 19.6\%, and 19.3\%, respectively. In terms of iRMSE, RIDERS (Depth Anything) achieves a reduction of 16.3\%, 15.9\%, and 15.6\% compared to the second-best method within the same ranges. Our method also demonstrates good performance on AbsRel, with the error reduced by 21.6\%, 21.7\%, and 21.6\% at 50m, 60m, and 70m, respectively, compared to the second-best method.

\begin{figure}[h]
    \centering
    \includegraphics[width=0.485\textwidth]{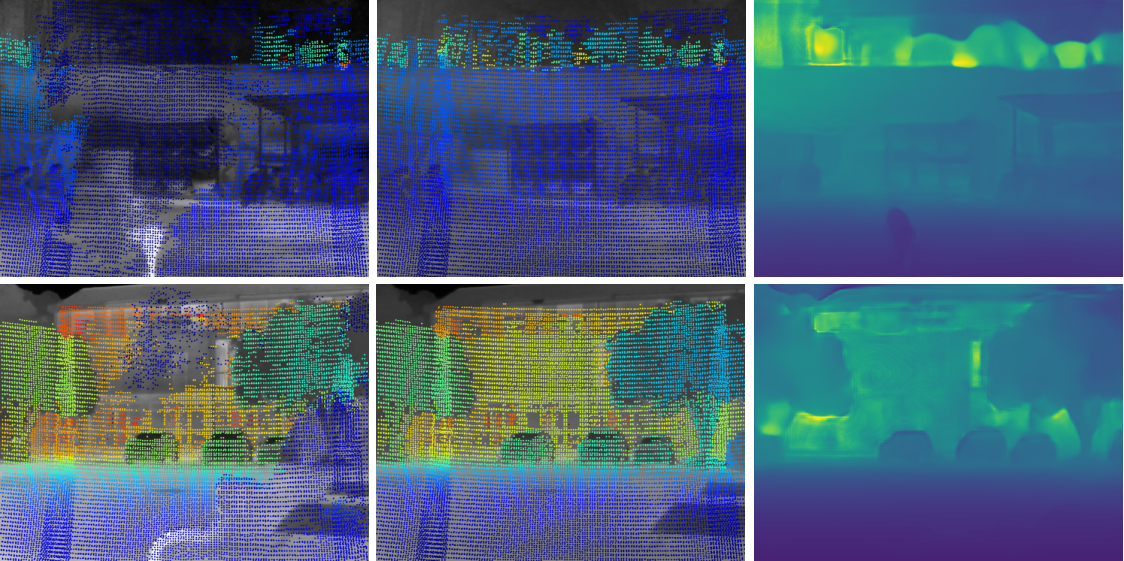}
    \captionsetup{font=small}
    \caption{\textbf{Showcased results on the smoke sequences of ZJU-Multispectrum dataset.} From left to right: LiDAR depth with smoke, its corresponding ground truth depth from the historical frame without smoke, and the depth estimation from RIDERS.}
    \label{fig:smoke-gt} 
\end{figure}

\begin{figure*}[t]
      \centering
      \includegraphics[width=1\textwidth]{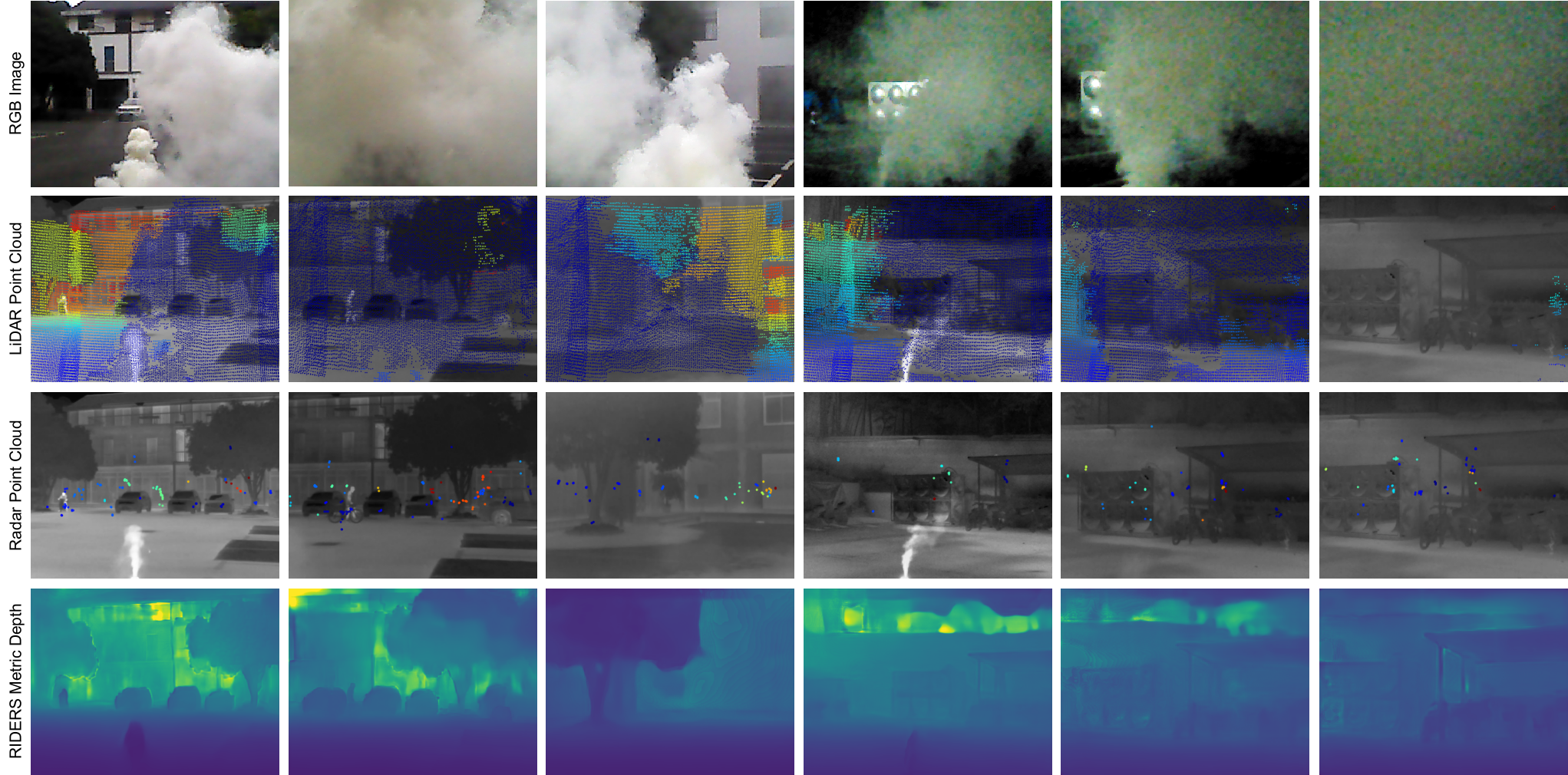}
      \caption{\textbf{Evaluations on ZJU-Multispectrum (smoke).} From top to bottom: RGB images in the visible spectrum, LiDAR point clouds, Radar point clouds, dense depth estimations of RIDERS. RIDERS performs remarkably well in scenarios where perception methods relying on visible light cameras and near-infrared LiDAR are severely impaired or face destructive challenges.}
      \label{fig:smoke}
      \vspace{0pt} 
\end{figure*}

\subsubsection{Nighttime or Smoke Scenarios} 
The primary objective of this work is to address the issue of perception (depth estimation) failure in adverse conditions. In this section, we evaluate our method in typical adverse scenarios under low-light and smoke conditions. For safety reasons, simulations of adverse scenarios were conducted in enclosed areas, resulting in a limited range of depth values. The evaluation of metric accuracy is calculated within a range of 50 meters. Results are shown in Tab.~\ref{ZJU smoke test}.

To evaluate the performance of our method under different lighting conditions, we conducted tests on both a clear-day sequence and a low-light sequence captured in the same scene. Notably, in clear-night conditions, our RIDERS method, benefiting from the low-light insensitivity of both the thermal camera and Radar, showcased accuracy comparable to that observed in the clear-day sequence.

To simulate challenging smoke-laden scenarios, we created artificial smoke with about 1 $\mu$m particle diameters. After that, we conducted tests when the sensor suite was kept stationary or moving in the presence of smoke.
In sequences affected by smoke, LiDAR measurements are proved to be unreliable due to interference from the smoke particles. To assess metric depth estimation under smoky conditions, we maintained a stationary position for the sensor suite. We then used LiDAR depths from previously captured stationary frames (absent of smoke) as a reference for evaluating frames that were impacted by smoke (see Fig.~\ref{fig:smoke-gt}). In this case, 446 synchronized Radar and thermal frames were available for metric evaluation. Our RIDERS exhibits high accuracy in both the nighttime and daytime sequences with smoke, shown in Tab.~\ref{ZJU smoke test}, and outperforms the other comparative methods \cite{lo2021depth, singh2023depth, li2024radarcam}.
To conduct thorough evaluations, we also subjected RIDERS to tests while the sensor suite was in motion, focusing on qualitative assessments across two smoke-impacted sequences. These tests further confirmed the robustness of RIDERS in handling smoke and nighttime conditions, as illustrated in Fig.~\ref{fig:smoke}. 

In our experiments,  the negligible impact of low light and atmospheric particulates on thermal cameras and millimeter-wave Radars ensures that RIDERS maintains its efficacy in scenarios severely compromising short-wave sensors such as LiDAR and RGB cameras. Consequently, RIDERS can deliver dependable metric depth estimates with consistent reliability across diverse conditions, including smoke-laden environments and both daytime and nighttime settings.

\subsection{Ablation}
Unless additional clarification is provided, the ablation studies referenced in this paper are conducted using the clear daytime sequences from the ZJU-Multispectrum dataset.

\subsubsection{Local Scale Refinement}
To validate the efficacy of our Scale Map Learner module, we evaluated the performance of the $\mathbf{\hat{d}}_{ga}$ without undergoing the local scale refinement described in Sec.~\ref{SML}. As shown in Tab.~\ref{dga test}, monocular depth prediction from LeReS \cite{yin2021learning}, ZoeDepth \cite{bhat2023zoedepth}, and Depth Anything \cite{yang2024depth} after global scale alignment with sparse Radar depth still exhibit poor accuracy, highlighting the need for further pixel-wise scale learning. Despite the lower metric accuracy of the $\mathbf{\hat{d}}_{ga}$ from Depth Anything, the inconsistency between local and global scales can be corrected through our SML. Our RIDERS (Any Depth) demonstrates the best performance overall compared to the comparisons.

\begin{table*}[th]
\centering
\caption{\textsc{Ablation of Local Scale Refinement} (mm).}
\label{dga test}
\resizebox{0.95\linewidth}{!}{
\begin{tabular}{c|c|c|c|c|c|c|c|c}
\hline\hline
{\textbf{Range}} & \textbf{Method} & \textbf{iMAE} $\downarrow$ & \textbf{iRMSE} $\downarrow$ & \textbf{MAE} $\downarrow$ & \textbf{RMSE} $\downarrow$ & \textbf{AbsRel} $\downarrow$& \textbf{SqRel} $\downarrow$ & $\mathbf{\delta_1}$ $\uparrow$ \\

\hline
\multirow{6}*{0-50m} 
{} & LeReS \cite{yin2021learning} & 26.420 & 32.966 & \textbf{6788.466} & \textbf{12056.420} & \textbf{0.381} & \textbf{7651.739} & 0.424 \\
{} & ZoeDepth \cite{bhat2023zoedepth} & \textbf{22.064} & \textbf{27.780} & 9384.187 & 20058.585 & 0.488 & 31493.616 & \textbf{0.494} \\
{} & Depth Anything \cite{yang2024depth} & 33.815 & 37.611 & 28224.456 & 50243.447 & 1.387 & 136253.932 & 0.112 \\
\cline{2-9}
{} & \textbf{RIDERS (LeReS)} & 10.423 & 15.755 & 2658.893 & 4765.552 & 0.147 & 1084.529 & 0.806 \\
{} & \textbf{RIDERS (ZoeDepth)} & 9.413 & 14.291 & 2576.012 & 4755.345 & 0.137 & 1081.774 & 0.826 \\
{} & \textbf{RIDERS (Depth Anything)} & \textbf{8.495} & \textbf{13.796} & \textbf{2423.499} & \textbf{4587.857} & \textbf{0.127} & \textbf{1025.969} & \textbf{0.848}\\

\hline\hline
\end{tabular}
}
\vspace{0pt} 
\end{table*}

\subsubsection{Scale Learning Strategy}

\begin{figure}[h]
      \centering
      \includegraphics[width=0.485\textwidth]{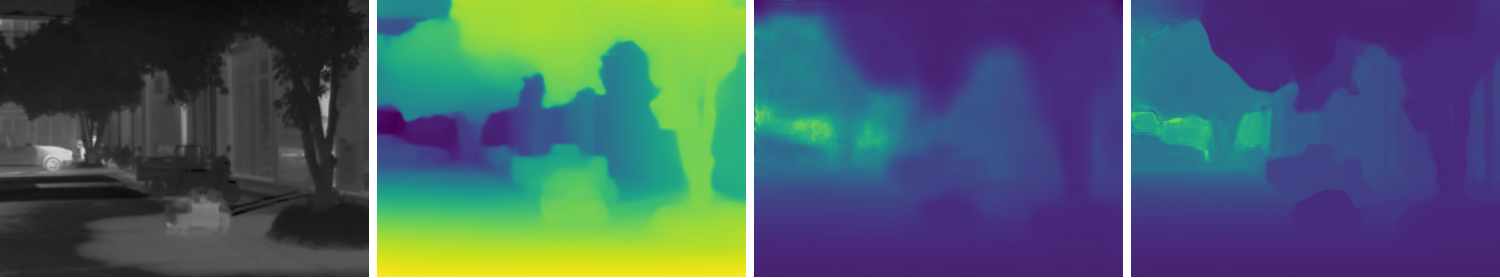}
      \captionsetup{font=small}
      \caption{\textbf{Ablation of the scale learning strategy.} From left to right: the input thermal image, monocular depth estimation from ZoeDepth \cite{bhat2023zoedepth}, metric depth estimation obtained through direct depth learning, and metric depth estimation obtained through scale learning.}
      \label{fig:ablation_scale}
      \vspace{0pt} 
\end{figure}

We executed a series of ablation experiments to examine the proposed approach of learning metric scale for monocular depth estimation. 
To provide a comparison, we trained the scale map learner (SML) with metric depth supervision, coercing it to output depth directly instead of scale.
Upon testing, directly learning depth resulted in blurry depth output and reduced convergence efficiency, as illustrated in Fig.~\ref{fig:ablation_scale}. The final evaluation results are also in the ``Direct Depth" rows of Tab.~\ref{ZJU clear test}. 
Notably, the strategy of directly learning depth produces significantly lower accuracy compared to learning metric scale for monocular depth.

The SML network takes input from monocular depth prediction, original thermal images, and augmented Radar depth.
Due to the sparsity and measurement noise of the Radar, pixel misalignment inevitably occurs when concatenating monocular depth prediction, original thermal images, and augmented Radar depth directly. This misalignment existing in the input of SML is eventually manifested as blurriness of the output depth estimation. If a network directly predicts depth maps with large propagated gradients during the training, this blurriness will be reflected as unclear object edges. However, if the network outputs a dense scale map with smaller gradients that represents local refinement, and the output will be multiplied onto $\hat{\mathbf{d}}_{ga}$, the impact of defective feature association blurriness and unclear edges will be significantly reduced. 
In other words, the preliminary depth from monocular prediction should not be viewed solely as a feature channel for learning. However, it should directly provide its most fundamental meaning -- depth, to our final depth estimation. Even if this preliminary monocular depth prediction does not have absolute metric accuracy, it can serve as a valid base to be explicitly corrected by the predicted dense scale map from SML.

\begin{figure}[h]
      \centering
      \includegraphics[width=0.485\textwidth]{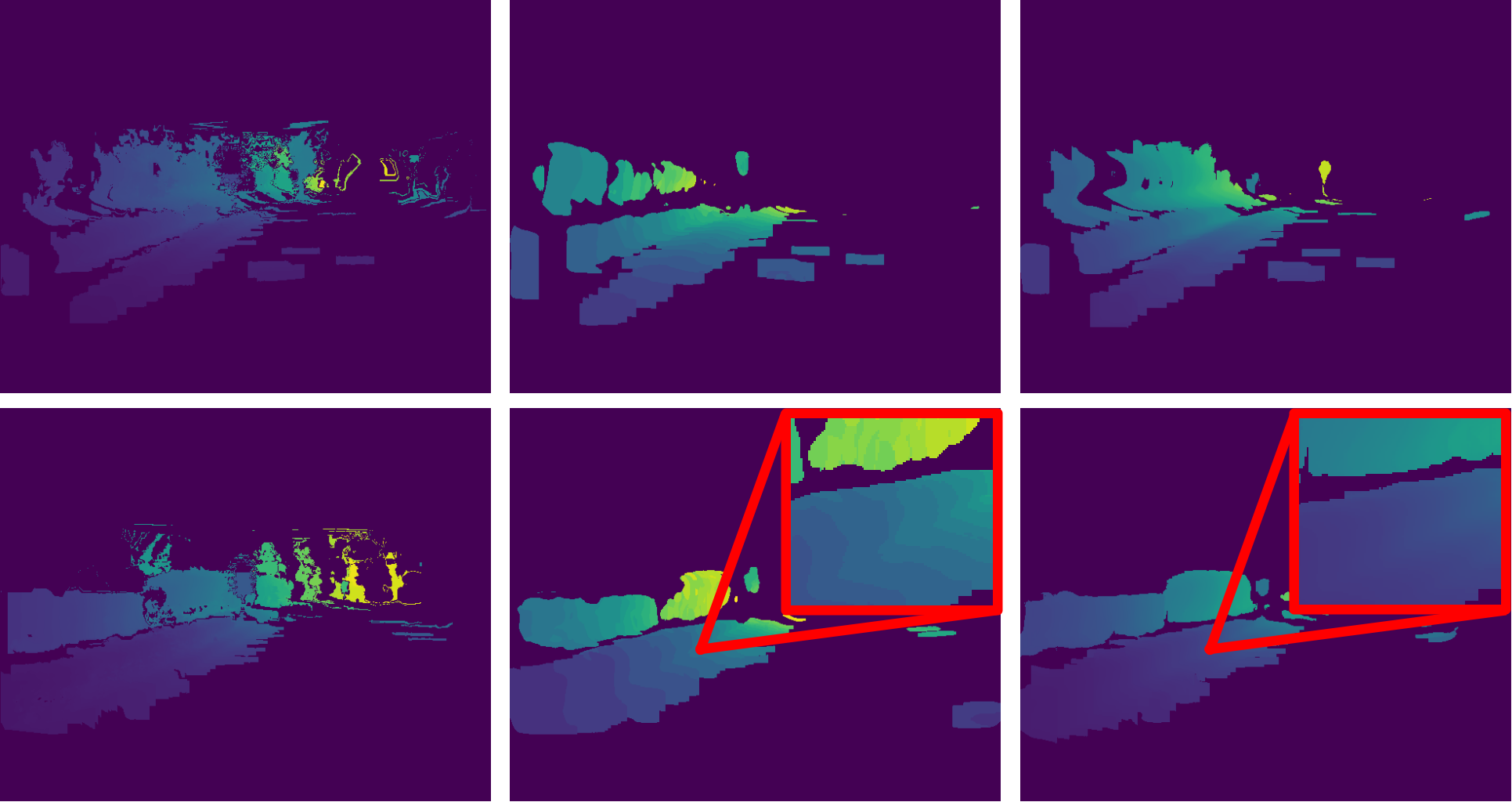}
      \captionsetup{font=small}
      \caption{\textbf{Ablation of Radar depth augmentation.} From left to right: the ground truth from the multi-patch supervision labels, the augmented depth from RC-vNet \cite{singh2023depth}, and the augmented depth from our proposed RC-Net. The detailed views of the quasi-dense depth are shown in the red box, demonstrating that our RC-Net can provide coherent and smooth depth estimation.}
      \label{fig:rcnet_compare}
      \vspace{0pt} 
\end{figure}

\begin{table*}[th]
\centering
\caption{\textsc{Ablation of Radar Depth Augmentation} (mm).}
\label{rcnet test}
\resizebox{0.72\linewidth}{!}{
\begin{tabular}{c|c|c|c|c|c}
\hline\hline
\textbf{Method} & \textbf{iMAE} $\downarrow$ & \textbf{iRMSE} $\downarrow$ & \textbf{MAE} $\downarrow$ & \textbf{RMSE} $\downarrow$ & \textbf{Output Pts} $\uparrow$ \\

\hline
RC-vNet \cite{singh2023depth} & 9.796 & 15.389 & 1191.473 & 2457.033 & 26692.283 \\
\hline
RC-vNet (Avg) \cite{singh2023depth} & 7.251 & 12.003 & 991.886 & 2288.220 & 26692.283 \\
\hline
\textbf{RC-Net (Avg)} & \textbf{6.289} & \textbf{9.820} & \textbf{921.780} & \textbf{1948.600} & \textbf{26791.745} \\

\hline\hline
\end{tabular}
}
\vspace{0pt} 
\end{table*}

\subsubsection{Radar Depth Augmentation}\label{sec:radar depth aug}

We conducted ablation experiments to analyze the role of RC-Net for Radar depth augmentation in the overall framework. As a comparison, we directly input the unenhanced sparse Radar points into SML for scale learning. The final dense depth estimation results are shown in the ``Sparse Input" rows in Tab.~\ref{ZJU clear test}, lagging behind in all metrics compared to the quasi-dense Radar input. Before the augmentation of RC-Net, only a few hundred Radar points were projected onto the image plane per frame. The augmented quasi-dense depth could provide a more extensive range of metric scales for SML. The ``Output Pts" in Tab.~\ref{ZJU clear test} denotes the number of points with valid depth after Radar depth augmentation.

To validate the effectiveness of our improvements to RC-vNet, we also test the benefits of the transformer module with attention mechanism and the quasi-dense depth calculation strategy on the NTU4DRadLM dataset.
Our RC-Net, compared to the RC-vNet, incorporates a transformer module into its network architecture. In terms of quasi-dense depth calculation, RC-vNet assigns the Radar depth with the highest confidence to the pixel, resulting in a blocky distribution and discontinuous depth changes in the augmented depth, deviating from the actual physical scene. In contrast, our method (``RC-Net (Avg)") employs a weighted average of the depths of multiple Radar points, making the quasi-dense depth closer to a continuous distribution in the real world, as shown in Fig.~\ref{fig:rcnet_compare}. 
Indeed, due to noise in Radar points and uncertainties in confidence inference, aggregating information from multiple points provides more context information, benefiting the depth augmentation.

We evaluated quasi-dense depth $\mathbf{\hat{d}}_q$ in the range of 0-50m using the interpolated dense ground truth $\mathbf{d}_{int}$.
As shown in Tab.~\ref{rcnet test}, the network architecture and the quasi-dense depth calculation strategy we employed significantly improved the accuracy of the augmented Radar depth. 
To independently validate the effectiveness of the transformer module, we modified the calculation strategy of RC-vNet to a weighted average, as shown in ``the RC-vNet (Avg)" row of Tab.~\ref{rcnet test}. It still falls behind our proposed RC-Net.

\section{Conclusion}
\label{sec:conclustion}

This paper introduces a novel method for robust metric depth estimation, combining millimeter-wave Radar point clouds with infrared thermal images. 
To address the performance degradation of existing depth estimation methods in challenging scenarios such as nighttime and smoke, we employed long-wave, low-light-impact Radar, and infrared thermal cameras as sensors to achieve robust depth estimation. We complement the metric depth estimation with a three-stage fusion process, which addresses the blurriness and significant noises for long-wave electromagnetic inputs. This process includes monocular depth prediction and global alignment, sparse Radar depth augmentation, and local scale refinement. Each stage leverages the respective advantages of multi-modal inputs effectively. Our approach based on learning local scale refinement avoids artifacts, noise, and blurring caused by the direct fusion of heterogeneous data encodings, showcasing good qualitative and quantitative results. 
In summary, we integrate a Radar and an infrared thermal camera for metric depth estimation in adverse scenarios. We also proposed the challenging ZJU-Multispectrum dataset and thoroughly tested our algorithm across diverse scenarios from multiple datasets. The conclusive findings of our study underscore that our pioneering multi-modal data fusion approach markedly improves upon the accuracy of current methodologies. This advancement addresses the existing void in Radar-Infrared fusion-based depth estimation techniques, significantly contributing to the field.
However, our proposed method still has limitations: the reference depths obtained from LiDAR, used for supervisory purposes, cover only a limited portion of the image, potentially complicating the learning-based scale refinement process. Furthermore, our approach is significantly dependent on preliminary monocular depth prediction, and the adaptability of models pre-trained on RGB datasets to thermal infrared imagery necessitates further enhancement. Future research endeavors could profitably explore unsupervised domain adaptation techniques to advance monocular depth prediction capabilities from infrared images.

{
\vspace{0.05cm}
\def\bibfont{\scriptsize}
\printbibliography
}

\end{document}